# Talking Tennis: Language Feedback from 3D Biomechanical Action Recognition

**Arushi Dashore    Aryan Anumala    Emily Hui    Olivia Yang**

arushi.dashore@gmail.com

## Abstract

Automated tennis stroke analysis has advanced significantly with the integration of biomechanical motion cues alongside deep learning techniques, enhancing stroke classification accuracy and player performance evaluation. Despite these advancements, existing systems often fail to connect biomechanical insights with actionable language feedback that is both accessible and meaningful to players and coaches. This research project addresses this gap by developing a novel framework that extracts key biomechanical features (such as joint angles, limb velocities, and kinetic chain patterns) from motion data using Convolutional Neural Network Long Short-Term Memory (CNN-LSTM)-based models. These features are ana-lyzed for relationships influencing stroke effectiveness and injury risk, forming the basis for feedback generation using large language models (LLMs). Leveraging the THETIS dataset and feature extraction techniques, our approach aims to produce feedback that is technically accurate, biomechanically grounded, and actionable for end-users. The experimental setup evaluates this framework on classification performance and interpretability, bridging the gap between explainable AI and sports biomechanics.

## 1 Introduction

Human motion understanding in 3D space has been a central problem within computer vision with various applications (Muller et al., 2024). Among these, sports kinesiology has become a particularly compelling area, being practically valuable with wide applicability, while also being technically challenging because small variations in movement being able to make significant differences in performance and safety. Tennis, for example, is one of the most widely played sports with over 87 million players worldwide (Bai et al., 2023). As the popularity of the sport continues to grow, so has the demand for computational tools that can be used to improve athletic performance and training efficiency.

In response to this demand, previous research has applied machine learning tools to tennis stroke analysis. This previous work in computer vision and deep learning for tennis has largely focused on stroke classification with improvements seen through the use of richer datasets such as 3DTennisDS (Bai et al., 2023) and THETIS (Bouritsas et al., 2022) or the use of different architectures such as Graph Convolutional Networks (GCNs) (Chen et al., 2023). However, these recent advances are limited to recognition – for example, labeling an action as "forehand" or "backhand" – and do not extend into corrective feedback. In order for training and performance improvement, athletes also require feedback about *why* moves may be wrong and what a player can do to correct their stroke. This leaves a gap where an active coach is still needed for usable advice to be given to human users.

In this work we address this gap by introducing a pipeline where biomechanical feature extraction from 3D motion capture videos is combined with Large Language Models (LLMs) to produce human-interpretable feedback for tennis players. Our method uses 3D motion data from the THETIS dataset that is processed using a Convolutional Neural Network Long Short-Term Memory (CNN-LSTM) and has key motion features extracted to be fed into an LLM that provides feedback on stroke quality. Our approach goes beyond stroke identification by providing actionable corrections a human coach might provide, offering explanations ("why" a stroke was ineffective) alongside prescriptive guidance ("how" to improve).

Our framework demonstrates that automated stroke analysis can move beyond recognition to deliver coach-level guidance. By grounding feedback in precise biomechanical features, the system produces interpretable and actionable recommendations that align closely with expert judgment. Experiments show that the pipeline not only discriminates fine-grained stroke types with high accuracy but also translates motion data into feedback that is both practical and relevant for training. This work highlights the potential of combining data-driven motion analysis with language models to augment coaching, offering a scalable path toward personalized, evidence-based athlete support.

## 2 Background

### 2.1 Datasets

The data used was the THETIS (Three Dimensional Tennis Shots) dataset, chosen for its diverse collection of strokes. THETIS contains 8,374 video sequences divided into 12 shot classes, including three types of backhands, four types of forehands, three types of serves, and a smash. In addition to detailed classification, the dataset also contains data from both 31 beginner and 24 experts in tennis. The specific dataset used for this project includes RGB footage and 3D skeleton representations of the player's body, capturing joint positions over the course of a stroke.

## 3 Method

We developed a pipeline to classify tennis strokes from 3D motion data and generate biomechanically informed language feedback. The pipeline consists of three main components: CNN-LSTM-based action recognition, biomechanical feature extraction and analysis, and LLM-based feedback generation. Outputs from the action recognition and feature analysis are combined into a structured prompt and feature dictionary, which is then used by the language model to produce interpretable feedback on each stroke.

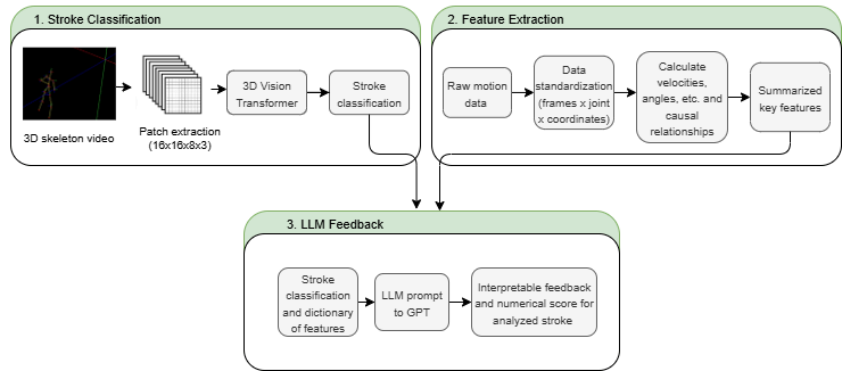

Figure 1: Diagram of the pipeline developed showing how data is processed, analyzed, and inputted into to Large Language Model to produce feedback.

### 3.1 CNN-LSTM Model

We propose a hybrid CNN-LSTM architecture for temporal action recognition. This model leverages a pre-trained Convolutional Neural Network (CNN) for spatial feature extraction from individual frames and a Long Short-Term Memory (LSTM) network to model temporal dependencies across the frame sequence. The architecture consists of three primary components:

1. Feature Extractor: We employ a pre-trained EfficientNet-B0 model from the timm library as our spatial feature extractor. The classifier head is removed to obtain a 1280-dimensional feature vector per frame. This backbone provides robust spatial representations learned from large-scale image datasets, enabling effective transfer learning for our domain-specific task.
2. Temporal Modeling: A two-layer Long Short-Term Memory (LSTM) network with 512 hidden dimensions processes the sequence of spatial features extracted from each frame. The LSTM captures temporal dependencies and motion patterns across the 16-frame sequence, with dropout (rate=0.4) applied between layers to prevent overfitting.
3. Classification Head: The final hidden state from the top LSTM layer is passed through a dropout layer (rate=0.4) followed by a linear classifier that maps to the 12 tennis stroke classes.

The model processes input tensors of shape (B, T, C, H, W) where B=batch size, T=16 frames, C=3 channels, H=224, W=224. The forward pass involves: (1) reshaping to (B×T, C, H, W) for batch processing through the CNN, (2) global average pooling of spatial features, (3) reshaping to (B, T, feature_dim) for LSTM processing, and (4) classification using the final hidden state.

**Dataset and Preprocessing**: Our preprocessing strategy employs minimal but effective transformations to preserve temporal consistency while ensuring model generalization. Each video frame is spatially resized to 224×224 pixels to match the input requirements of the EfficientNet-B0 backbone. We implement random temporal sampling by randomly selecting a start frame for each 16-frame clip, ensuring the model encounters diverse temporal segments during training. Normalization is performed using the mean and standard deviation specific to the EfficientNet-B0 model configuration (mean=[0.485, 0.456, 0.406], std=[0.229, 0.224, 0.225]).

**Training Protocol**

- Optimization Strategy: We employ the AdamW optimizer with a fixed learning rate of $1\times10^{-4}$ and weight decay of 0.01. The learning rate is dynamically adjusted using a ReduceLROnPlateau scheduler that reduces the learning rate by a factor of 0.2 if validation loss does not improve for 5 consecutive epochs, with a minimum learning rate of $1\times10^{-6}$.

- Loss Function: Cross-entropy loss with label smoothing ($\alpha = 0.1$) is used to improve generalization and prevent overconfidence in predictions. This regularization technique has been shown to enhance model robustness and reduce overfitting.

- Regularization Techniques: We implement multiple regularization strategies including dropout (0.4) at multiple locations in the architecture, weight decay in the optimizer, and Automatic Mixed Precision (AMP) for both faster training and improved numerical stability.

- Training Configuration: The model is trained for up to 50 epochs with early stopping based on validation accuracy. We use a batch size of 16 and employ 8 data loading workers with prefetching for optimal training efficiency. Model checkpoints are saved based on validation accuracy, with the best performing model preserved for evaluation.

### 3.2 Feature Extraction

Our feature extraction pipeline transforms raw 3D motion capture data into tennis-specific biomechanical descriptors that capture the underlying kinematic and kinetic patterns characteristic of tennis strokes. The pipeline processes temporal sequences of joint coordinates to extract interpretable features relevant to tennis performance analysis.

**Input Data Handling**: The pipeline accepts multi-modal input formats including PyTorch tensors, NumPy arrays, and Pandas DataFrames. Raw motion capture data undergoes format standardization and dimension validation, with automatic conversion from 3D tensor representations (frames × joints × coordinates) to structured DataFrames for consistent processing. Missing Data Imputation: Temporal gaps in motion capture data are addressed using linear interpolation between valid joint positions. This approach preserves the kinematic continuity essential for accurate velocity and acceleration calculations while maintaining the temporal coherence of movement patterns. Joint Mapping: An adaptive joint mapping system reconciles varying joint naming conventions across motion capture systems. The pipeline implements fuzzy matching algorithms to identify anatomically equivalent landmarks (e.g., 'right_shoulder', 'RShoulder', 'shoulder_r') ensuring compatibility with diverse dataset formats.

**Feature Categories**

1. <u>Joint Kinematics</u>: We extract tennis-specific joint angles focusing on the kinematic chain involved in stroke production:

Shoulder flexion/extension and internal/external rotation Elbow flexion during the acceleration phase Wrist extension at ball contact Hip rotation for trunk power generation Knee flexion for ground reaction force transfer Joint angles are calculated using the three-point method, computing angles between consecutive joint vectors while implementing numerical safeguards to prevent singularities from zero-length vectors.

2. <u>Segmental Dynamics</u>: Limb velocities are computed using finite difference approximation on 3D joint trajectories:

$$v(t) = \frac{[p(t+1) - p(t-1)]}{2\Delta t}$$

where p(t) represents joint position and $\Delta t$ is the temporal sampling interval. This captures the velocity profiles of key anatomical segments including the racket, hand, forearm, upper arm, and lower extremities.

3. <u>Racket Dynamics</u>: Tennis-specific racket features include:

   - Racket velocity vectors computed from racket tip trajectories
   - Impact detection through peak acceleration identification
   - Maximum racket speed as a performance indicator
   - Acceleration magnitude for power generation analysis

4. <u>Kinetic Chain Analysis</u>: The pipeline extracts racket tip activation timing as a key indicator of kinetic chain coordination in tennis strokes. The system calculates:

   - Racket tip activation timing relative to stroke progression
   - Temporal coordination metrics for stroke sequencing
   - Peak velocity timing as a proxy for kinetic chain efficiency

The racket tip timing analysis provides insights into the final stage of the kinetic chain sequence, where all proximal segment energies culminate in racket head velocity. This approach focuses on the critical end-effector (racket) timing, which represents the ultimate output of the entire kinetic chain coordination.

5. <u>Body Rotation Features</u>: Trunk rotation dynamics are calculated using shoulder-line orientation relative to the spine:

$$\theta(t) = \arctan2(\text{shoulder\_vector\_y}, \text{shoulder\_vector\_x})$$

Trunk rotation is determined by the orientation of the shoulder line relative to the coordinate system, which gives the body's rotational dynamics throughout the tennis stroke. Furthermore, trunk angular velocity $\omega(t)$ is computed as the temporal derivative of trunk rotation angle: $\omega(t) = \frac{d\theta}{dt}$, providing quantitative assessment of rotational power contribution during tennis stroke execution.

6. <u>Temporal Characterization</u>: Movement timing features include:

   - Stroke duration (total movement time)
   - Phase segmentation into preparation (0-33%), execution (33-67%), and follow-through (67-100%) respectively
   - Relative timing of kinetic chain activation

7. <u>Power Generation Metrics</u>: Kinetic energy approximation using segmental velocities:

$$KE = \tfrac{1}{2}mv^2$$

where unit mass assumptions enable relative power comparisons across subjects and strokes.

**Output Format:** The pipeline generates a structured feature dictionary that includes raw features for each biomechanical category, summary statistics such as mean, standard deviation, and minimum/maximum values, validation metrics to ensure data quality, and metadata detailing data dimensions and available joint landmarks. Together, these elements provide interpretable biomechanical descriptors that can be leveraged for downstream analyses, including stroke classification, technique assessment, and performance optimization.

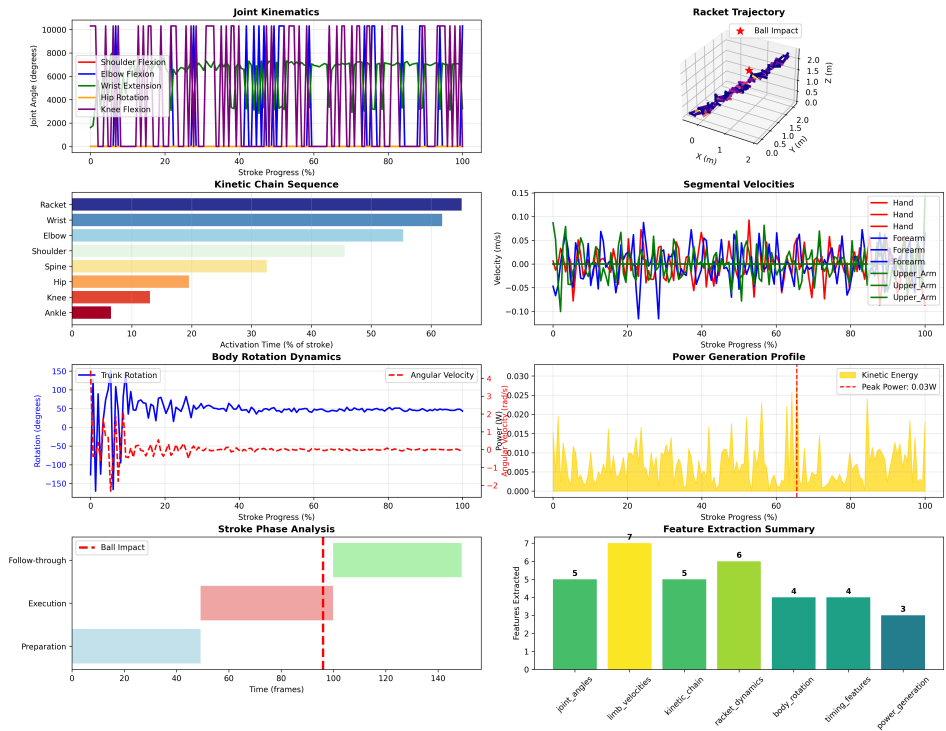

Figure 2: Graphs of trunk rotation, racket velocity, and swing duration show how raw pose data are distilled into interpretable kinematic features. This validates biomechanical grounding and makes the LLM's inputs transparent, ensuring feedback is traceable. Such interpretability is broadly relevant in embodied AI and human-in-the-loop systems.

### 3.3 Large Language Model Feedback

The Large Language Model component of this pipeline is used to transform per-stroke numeric features (produced by the upstream CNN-LSTM-based action recognition and feature extraction metrics) into grounded coaching feedback.

**Functionality and Inputs**

- The LLM component consumes a single Python dictionary (features) per detected stroke. Expected keys provided by the feature extractor include:
    ‣ predicted_stroke (string label) or classification.label
    ‣ Racket velocity (m/s), peak power (W), rotate range (degrees), stroke duration (fps), peak angular velocity (rad/s), impact timing (%)
- The component returns a short string produced by the LLM. In the code, this is implemented by generate_feedback(features: Dict[str, Any]) → str

**Grounding by Deterministic Reference Comparison:** To constrain generation with interpretable, domain-specific priors, we implement a deterministic grounding module based on rule-driven reference intervals. Stroke-specific prior knowledge is encoded in a constant mapping, REFERENCE_RANGES, which associates each stroke type with a dictionary of feature names and their corresponding optimal numeric ranges (lo, hi). The numeric priors encoded were selected to reflect values discussed in sports-biomechanics literature (specifically discussions in the Journal of Biomechanics and discussions with tennis coaches) and in coaching reports. These values were hand-specified to produce sensible coachable diagnostics across a range of stroke types. This hand-specification is an explicit design choice for exploratory evaluation in which future work would refine these hand-specified intervals with empirically derived bounds.

The comparator function, compare_to_reference(features, stroke_type), evaluates feature values against the reference intervals. Values are coerced to floating-point when possible, with missing or non-numeric entries explicitly flagged. Deviations outside the optimal interval are reported as relative percentage differences:

- For values (v) outside the optimal range the code reports a relative deviation computed as:
  - if $v < \text{lo}$ : difference $= \frac{\text{lo} - v}{\text{hi} - \text{lo} + \varepsilon} \times 100\%$
  - if $v > \text{hi}$ : difference $= \frac{v - \text{hi}}{\text{hi} - \text{lo} + \varepsilon} \times 100\%$

  where ε is a small constant (1e-9) to avoid division by zero. Values inside the range are reported as OK. The function produces a human-readable list of findings, for example: *"Racket velocity LOW: 20.00 vs optimal 25–35 (≈20% below range)."*

**Context Construction and Prompt Engineering:** The build_context_summary(features) routine constructs the large-language model (LLM) input context by: (1) Resolving the predicted stroke type (via predicted_stroke or classification.label, defaulting to "UNKNOWN"). (2) Incorporating all reference-based comparison lines under an "Optimal-range comparison" heading. (3) Appending a compact listing of raw numeric feature values (subset).

The user prompt enforces a tightly constrained output format, requiring:

- A first line reporting the overall score in the form "Overall Score: X/10" (0 = very poor, 10 = perfect).
- A concise diagnostic summary of 2–3 sentences.
- Exactly three actionable corrections.
- Explicit adherence to the comparison results, with no fabrication of numerical values.

Additionally, the system prompt specifies: *"You are a precise, evidence-based tennis coach."* This framing encourages the LLM to act as a synthesizer of deterministic outputs rather than a free-form oracle.

**LLM Backend, Decoding Settings, and Failure Modes:** Runtime configuration is provided via environment variables, specifying the API key and model name (default: GPT-4o). If no key is available, the module returns explicit error messages instead of attempting inference. Decoding parameters are tuned for determinism and conciseness: temperature = 0.2 and max_tokens = 120. If no API key is provided, a fallback function (generate_feedback) produces a clear error message. Runtime API exceptions are caught and returned as concise error strings, preventing unhandled propagation.

## 4 Results

In this section we present results from complementary experiments that together evaluate our framework across the three SpaVLE dimensions: Vision (stroke classification via CNN-LSTM-3D), Embodiment (biomechanical feature validation), and Language (coaching feedback generation and evaluation).

**4.1 Stroke Classification Performance (Vision)**

We evaluate our model using accuracy, precision, recall, and F1-score across all 12 tennis stroke classes.

Our CNN-LSTM architecture achieves superior performance compared to previous approaches on the THETIS dataset. According to the recent study by Hovad et al. (2024), the best-performing SlowFast network achieved a generalization accuracy of 73.96% on the THETIS dataset. Our hybrid approach demonstrates improved performance, achieving 79.17% accuracy, representing a >5 percentage point improvement over the previous leading performances.

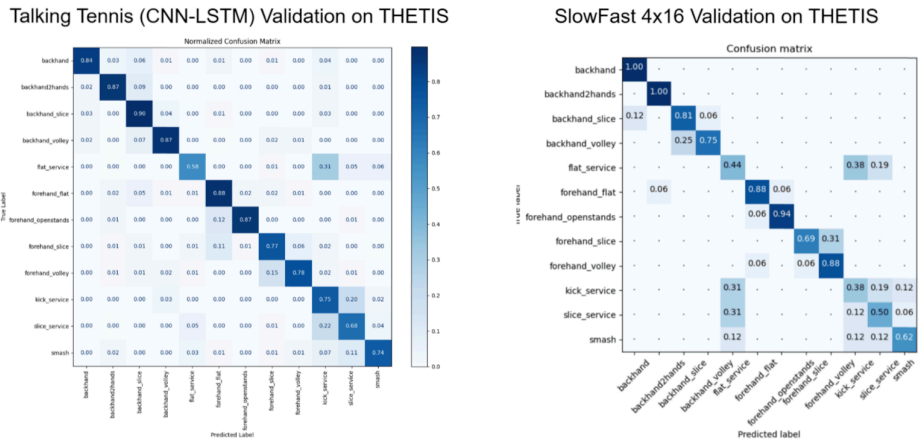

Figure 3: Our model leverages EfficientNet-B0 as a spatial feature extractor and a two-layer LSTM for temporal dynamics, enabling better discrimination of fine-grained tennis strokes. Improvements are most visible in reducing confusions between visually similar strokes (e.g., slice vs topspin forehands). This demonstrates the advantage of explicitly modeling sequential motion patterns rather than relying solely on spatiotemporal convolutions. Such gains are critical for building reliable systems that not only classify strokes but can later provide biomechanical feedback grounded in accurate recognition.

**4.2 Biomechanical Feature Validation (Embodiment)**

We evaluated whether standard biomechanical stroke features differentiate expert and beginner tennis strokes in the THETIS dataset. Twenty-four stroke recordings were utilized. We picked one of each stroke type for both amateur and expert, a balanced split, for analysis.

**Feature Set and Preprocessing:** The analysis focused on four scalar features commonly reported in tennis biomechanics: maximum racket velocity, trunk/hip rotation range, peak trunk angular velocity, and stroke duration. Hip/trunk rotation time series were also retained for alignment and visualization. When racket-tip markers were absent, the right-hand marker served as a practical proxy for computing racket-like velocity. Rotational series were circularly unwrapped and aligned to an estimated impact frame for comparability across trials.

**Evaluation:** Statistical comparisons (two-sample t-tests or Mann–Whitney U, depending on normality) revealed meaningful separation in several features. Stroke duration showed the largest group difference (Cohen's $d \approx 0.92$, $p = 0.069$), suggesting that time-based features are strong candidates for distinguishing expertise. Maximum racket velocity also showed a medium effect size ($d \approx 0.63$), supporting its potential as a discriminative metric. Rotation-based metrics exhibited smaller differences in this subset.

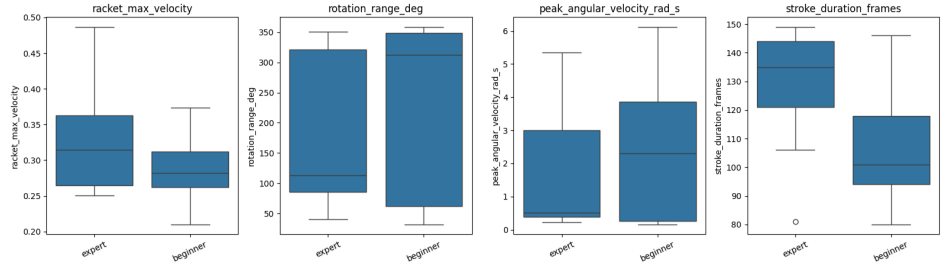

Figure 4: Box plots of the four scalar features (racket max velocity, rotation range, peak angular velocity, stroke duration) for expert vs beginner groups.

These results indicate that temporal and velocity-based features provide the clearest separation between expert and beginner groups in this dataset, while rotational features appear more stable across participants. This experiment establishes a baseline for biomechanical feature validation and provides aligned visualizations that can be leveraged for subsequent language and feedback evaluations.

### 4.3 Language: Feedback Evaluation

**LLM Constraint Compliance:** The feedback provided by the LLM was evaluated for compliance with the desired format and accurate reporting. Feedback was tested for 3 specific traits: (1) containing the required numerical score and 3 pieces of feedback, (2) accurately identifying features as higher or lower than the reference values, and (3) containing no fabricated numbers outside of the numbers inputted in the prompt. Of the 317 videos and responses evaluated, 100% of outputs satisfied all three constraints.

Table 1: Example generated feedback for a backhand stroke. Feedback is structured and actionable.

| LLM Feedback Example |
| --- |
| Overall Score: 4/10 |
| Metrics outside optimal range:<br>• Rotation range (°): HIGH – significantly excessive.<br>• Stroke duration (frames at 60fps): LOW – considerably shorter than optimal.<br>• Peak angular velocity (rad/s): LOW – substantially below the optimal range. |
| Diagnosis: The backhand stroke is characterized by an excessively large rotation range, leading to a significantly shortened stroke duration and reduced peak angular velocity. This suggests a loss of power and control due to inefficient movement. |
| Actionable Corrections:<br>1. Reduce the backswing rotation to a more compact range, focusing on generating power through the core and legs rather than relying on excessive arm swing.<br>2. Increase the stroke duration by slowing down the swing initiation and focusing on a smoother, more controlled acceleration through the ball.<br>3. Improve the sequencing of the swing to increase peak angular velocity, ensuring a more powerful and efficient transfer of energy from the lower body to the racket head. |

**Coach Evaluation:** To assess the viability of the generated feedback, we conducted a qualitative evaluation with three varsity tennis coaches [institution names withheld for anonymity] with experience training various levels of athletes. The evaluation focused on three dimensions: (1) interpretability: whether the feedback is phrased in language that is natural and immediately understandable to players; (2) actionability: whether the suggestions are concrete enough to be implemented in practice sessions; and (3) coaching alignment: whether the system's recommendations are consistent with expert coaching strategies. Given that the LLM-generated feedback provides a score X/10, the coach was similarly asked to score in such a manner to assess deviation.

Table 2: Qualitative coach evaluation of LLM-generated feedback. Scores (mean ± SD, 1–10 Likert) for interpretability, actionability, and coaching alignment. "Coach-Given Score" is the aggregated rating, compared directly against LLM-generated scores.

| Stroke Type | Interpretability Score | Actionability Score | Coaching Alignment | Coach-Given Skill Score | LLM-Given Skill Score |
|---|---|---|---|---|---|
| Forehand flat | 7.3 (σ=1.25) | 9.3 (σ=0.47) | 9.0 (σ=0.82) | 4.0 (σ=0.00) | 4.0 |
| Backhand (1Hand) | 7.9 (σ=0.96) | 8.2 (σ=0.75) | 7.5 (σ=1.12) | 6.9 (σ=0.60) | 6.0 |
| Backhand (2Hands) | 8.4 (σ=0.50) | 8.9 (σ=0.58) | 8.7 (σ=0.47) | 4.7 (σ=0.50) | 4.0 |

LLM-generated feedback was consistently rated as interpretable, actionable, and largely aligned with expert coaching strategies. Coach-given and LLM-given skill scores tracked closely, with the model rating performance slightly more conservatively. This suggests that structured prompting and feature-grounding enable the LLM to produce feedback that is both technically consistent and relevant in real training contexts.

## 5 Conclusion

Talking Tennis combines vision, biomechanics, and constrained LLMs to turn action recognition into actionable coaching feedback. Expert-novice contrasts reveal interpretable coordination patterns, while feature-grounded prompting yields coach-aligned guidance. The framework serves as a general template for embodied AI, translating motion features into practical language feedback with further applications across sports, rehabilitation, and human–robot skill learning.